\documentclass[lettersize,journal]{IEEEtran}
\usepackage[T1]{fontenc}
\usepackage{amsmath,amsfonts}
\usepackage{algorithmic}
\usepackage{array}
\usepackage[caption=false,font=normalsize,labelfont=sf,textfont=sf]{subfig}
\usepackage{textcomp}
\usepackage{stfloats}
\usepackage{url}
\usepackage{verbatim}
\usepackage{graphicx}
\usepackage{multirow}
\def\BibTeX{{\rm B\kern-.05em{\sc i\kern-.025em b}\kern-.08em
    T\kern-.1667em\lower.7ex\hbox{E}\kern-.125emX}}
\usepackage{balance}
\begin{document}
\title{Survey on Controlable Image Synthesis with Deep Learning}

\author{
  \IEEEauthorblockN{Shixiong Zhang}
  \thanks{Shixiong Zhang is with the School of Automation Engineering, University of Electronic Science and Technology of China, Chengdu, Sichuan, China (email: 202221060818@std.uestc.edu.cn).}
  \and
  \IEEEauthorblockN{Jiao Li}
  \thanks{Jiao Li is with the College of Information Engineering, Sichuan Agricultural University, Chengdu, Sichuan, China (email: 202005852@stu.sicau.edu.cn).}
  \and
  \IEEEauthorblockN{Lu Yang}
  \thanks{Lu Yang is with the School of Automation Engineering, University of Electronic Science and Technology of China, Chengdu, Sichuan, China (email: yanglu@uestc.edu.cn). Corresponding Author. Member, IEEE.}
}

\markboth{Journal of \LaTeX\ Class Files,~Vol.~18, No.~9, September~2020}%
{How to Use the IEEEtran \LaTeX \ Templates}

\maketitle

\begin{abstract}
	Image synthesis has attracted emerging research interests in academic and industry communities. Deep learning technologies especially the generative models greatly inspired controllable image synthesis approaches and applications, which aim to generate particular visual contents with latent prompts. In order to further investigate low-level controllable image synthesis problem which is crucial for fine image rendering and editing tasks, we present a survey of some recent works on 3D controllable image synthesis using deep learning. We first introduce the datasets and evaluation indicators for 3D controllable image synthesis. Then, we review the state-of-the-art research for geometrically controllable image synthesis in two aspects: 1) Viewpoint/pose-controllable image synthesis; 2) Structure/shape-controllable image synthesis. Furthermore, the photometrically controllable image synthesis approaches are also reviewed for 3D re-lighting researches. While the emphasis is on 3D controllable image synthesis algorithms, the related applications, products and resources are also briefly summarized for practitioners.
\end{abstract}


\section{Introduction}
\label{sec1}
\IEEEPARstart{A}{rtificial} Intelligence Generated Content (AIGC) is the term for digital media produced by machine learning methods, such as ChatGPT and stable diffusion\cite{Rombach_2022_CVPR}, which are currently popular\cite{cao2023comprehensive}.  AIGC has various applications in domains such as entertainment, education, marketing, and research\cite{Bommasani2021FoundationModels}.  Image synthesis is a subcategory of AIGC that involves generating realistic or stylized images from textual inputs, sketches, or other images\cite{zhang2023adding}. Image synthesis can also perform various tasks such as inpainting, semantic scene synthesis, super-resolution, and unconditional image generation\cite{Rombach_2022_CVPR,wang2021realesrgan,DDPM,GAN}.

Image synthesis can be classified into two types based on controllability: unconditional and conditional \cite{huang2018introduction}. Conditional image synthesis can be further divided into three levels of control: high, medium, and low. High-level control refers to the image content such as category, medium-level control refers to the image background and other aspects, and low-level control refers to manipulating the image based on the underlying principles of traditional computer vision \cite{mirza2014conditional,Gatys2016,Agarwal2009}.

Conventional 3D image synthesis techniques face challenges in handling intricate details and patterns that vary across different objects \cite{yang2010}. Deep learning methods can better model the variations in shape, texture, and illumination of 3D objects\cite{ZHENG2022}. The field of deep learning-based image synthesis has made remarkable progress in recent years, aided by the availability of more open source datasets \cite{ImageNet,LAION-5B,WikiArt}. Various image synthesis methods have emerged, such as generative adversarial network (GAN) \cite{GAN}, diffusion model (DM) \cite{DDPM}, and neural radiance field (NeRF) \cite{NeRF}. These methods differ in their levels of controllability: GAN and DM are suitable for high-level or medium-level controllable image synthesis, while NeRF is suitable for low-level controllable image synthesis.

\begin{figure*}[b]
	\centering
		\subfloat[Original image]{\includegraphics[width = 0.2\textwidth]{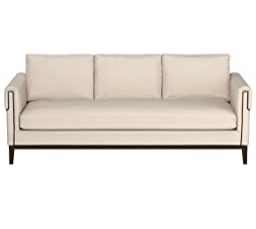}}
		\hfill
		\subfloat[Image after changing viewpoint]{\includegraphics[width = 0.2\textwidth]{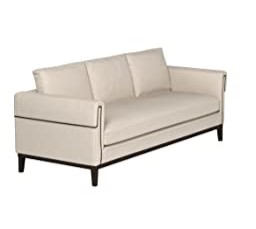}}
		\hfill
		\subfloat[Original image]{\includegraphics[width = 0.2\textwidth]{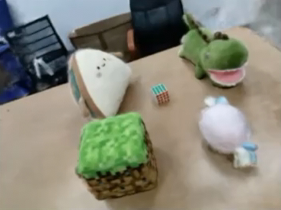}}
		\hfill
		\subfloat[Image after rotating object]{\includegraphics[width = 0.2\textwidth]{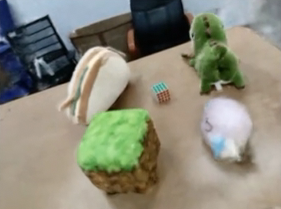}}
		\caption{Pose manipulation example.Subfigure (a) and (b) correspond to the content of Global Pose realization in Section \ref{sec3}, which come from \cite{collins2022abo}. Subfigure (c) and (d) correspond to the content of Local Pose realization in Section \ref{sec3}, which come from \cite{yang2021objectnerf}}
		\label{fig:1}
	\end{figure*}

\begin{figure*}[bht]
	\centering
		\subfloat[Original image]{\includegraphics[width = 0.2\textwidth]{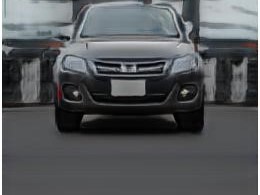}}
		\hfill
		\subfloat[Image after changing the depth of car]{\includegraphics[width = 0.2\textwidth]{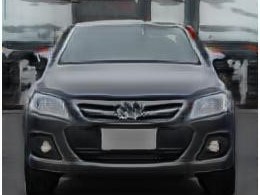}}
		\hfill
		\subfloat[Original image]{\includegraphics[width = 0.2\textwidth]{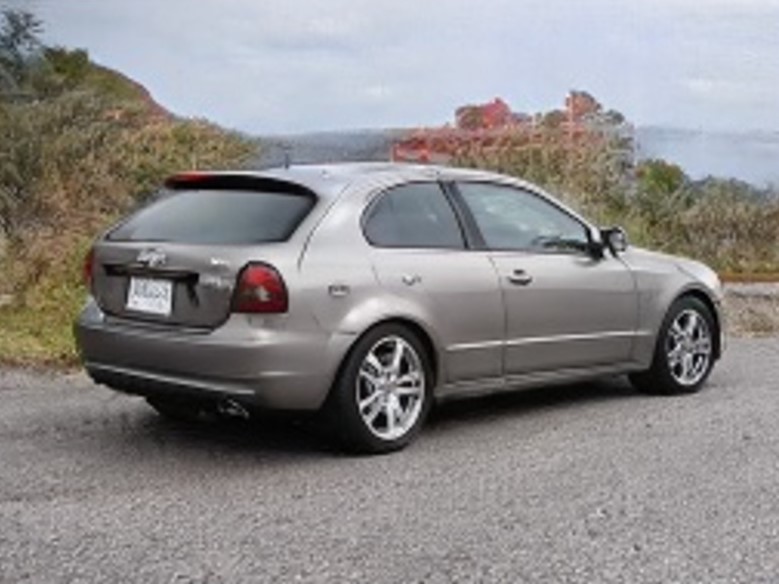}}
		\hfill
		\subfloat[Image after changing the color of car]{\includegraphics[width = 0.2\textwidth]{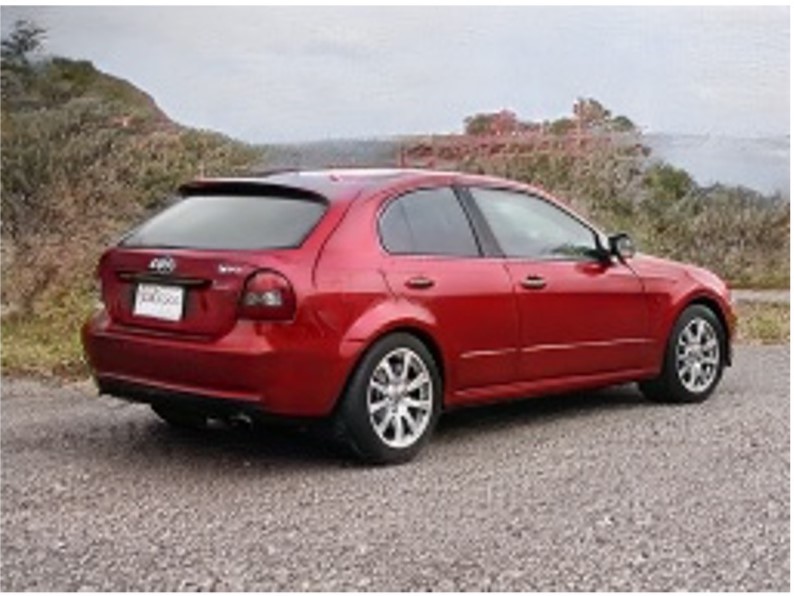}}
		\caption{Structure manipulation example.Subfigure (a) and (b) correspond to the content of Global Structure realization in Section \ref{sec4}, which come from \cite{Niemeyer2020GIRAFFE}. Subfigure (c) and (d) correspond to the content of Local Structure realization in Section \ref{sec4}, which come from \cite{zhu2023linkgan}}
		\label{fig:2}
\end{figure*}

\begin{figure*}[bht]
	\centering
		\subfloat[Original image]{\includegraphics[width = 0.2\textwidth]{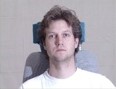}}
		\hfill
		\subfloat[Image after changing light source]{\includegraphics[width = 0.2\textwidth]{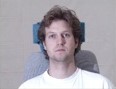}}
		\hfill
		\subfloat[Original image]{\includegraphics[width = 0.2\textwidth]{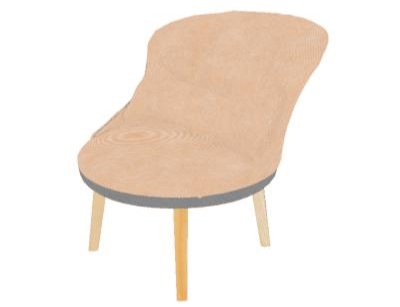}}
		\hfill
		\subfloat[Image after changing roughness]{\includegraphics[width = 0.2\textwidth]{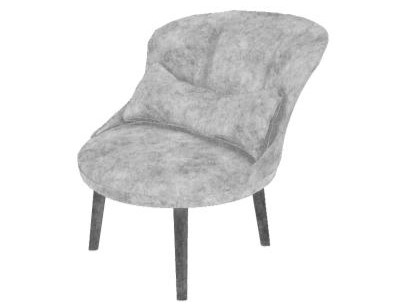}}
		\caption{Illumination manipulation example.Subfigure (a) and (b) correspond to the content of Global Illumination realization in Section \ref{sec5}, which come from \cite{Multi-PIE2008}. Subfigure (c) and (d) correspond to the content of Local Illumination realization in Section \ref{sec5}, which come from \cite{boss2021nerd}}
		\label{fig:3}
\end{figure*}

Low-level controllable image synthesis can be categorized into geometric and illumination control. Geometric control involves manipulating the pose and structure of the scene, where the pose can refer to either the camera or the object, while the structure can refer to either the global shape (using depth maps, point clouds, or other 3D representations) or the local attributes (such as size, shape, color, etc.) of the object. Illumination control involves manipulating the light source and the material properties of the object. Refer to Fig. \ref{fig:1}, \ref{fig:2}, and \ref{fig:3} for an example figure.

Several surveys have attempted to cover the state-of-the-art techniques and applications in image synthesis. However, most of these surveys have become obsolete due to the rapid development of the field \cite{huang2018introduction}, or have focused on the high-level and medium-level aspects of image synthesis, while ignoring the low-level aspects \cite{controlable_overview}. Furthermore, most of these surveys have adopted a methodological perspective, which is useful for researchers who want to understand the underlying principles and algorithms of image synthesis, but not for practitioners who want to apply image synthesis techniques to solve specific problems in various domains \cite{controlable_overview, Tsirikoglou2020ASO}. This paper provides a task-oriented review of low-level controllable image synthesis, excluding human subjects\cite{haas2022controllable,zhang2022training,ko20233d,Yang2023}.

\begin{figure}[t]
	\centering
	\includegraphics[width = 0.48\textwidth]{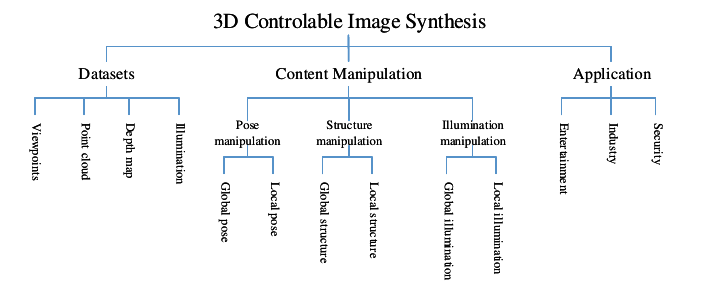}
	\caption{The structure diagram of this paper}
	\label{fig:8}
\end{figure}

This review offers a comprehensive overview of the state-of-the-art deep learning methods for 3D controllable image synthesis.     In Section \ref{sec2},  we begin by introducing the common data sets and evaluation indicators for this task. For the data set part, we divide it by its content.       In Section \ref{sec3} to \ref{sec5}, we survey the control methods based on pose, structure and illumination, and divided each part into global and local controls.       In Section \ref{sec6}, we discuss some of the current applications of 3D controllable image synthesis based on deep learning.  Finally, Section \ref{sec7} concludes this paper. The overview of the surveyed 3D controlable image synthesis is shown in Fig. \ref{fig:8}.In the following sections, we will review common data sets and evaluation indicators in detail.

\section{Data Sets and Evaluation Indicators for 3D Controlable Image Synthesis}
\label{sec2}

One of the key challenges in 3D controllable image synthesis is to evaluate the quality and diversity of the generated images. Different data sets and metrics have been proposed to measure various aspects of 3D controllable image synthesis, such as realism, consistency, fidelity, and controllability. In this section, we will introduce some of the commonly used data sets and metrics for 3D controllable image synthesis, and discuss their advantages and limitations.

\subsection{Data Sets}

3D image synthesis is the task of generating realistic images of 3D objects from different viewpoints. This task requires a large amount of training data that can capture the shape, texture, lighting and pose variations of 3D objects. Several datasets have been proposed for this purpose, each with its own advantages and limitations. Some of the data sets are:

\begin{itemize}
 	\item ABO is a synthetic data set that contains 3D shapes generated by assembling basic objects (ABOs) such as cubes, spheres, cylinders, and cones. It has 10 categories and 1000 shapes per category. ABO is useful for tasks such as shape abstraction, decomposition, and generation. However, ABO is also limited by its synthetic nature, its small number of categories and instances, and its lack of realistic lighting and occlusion\cite{collins2022abo}.
	\item Clevr3D is a synthetic data set that contains 3D scenes composed of simple geometric shapes with various attributes such as color, size, and material. It also provides natural language descriptions and questions for each scene. Clevr3D is useful for tasks such as scene understanding, reasoning, and captioning. However, Clevr3D is also limited by its synthetic nature, its simple scene composition, and its lack of realistic textures and backgrounds\cite{yan2021clevr3d}.
	\item ScanNet is an RGB-D video data set that contains 2.5 million views in more than 1500 scans of indoor scenes. It provides annotations such as camera poses, surface reconstructions, and instance-level semantic segmentations. ScanNet is useful for tasks such as semantic segmentation, object detection, and pose estimation.ScanNet is also limited by its incomplete coverage (due to scanning difficulties), its inconsistent labeling (due to human errors), and its lack of fine-grained details (such as object parts)\cite{dai2017scannet}.
	\item RealEstate10K is a data set for view synthesis that contains camera poses corresponding to 10 million frames derived from about 80,000 video clips gathered from YouTube videos. The data set also provides links to download the original videos. RealEstate10K is a large-scale and diverse data set that covers various types of scenes, such as houses, apartments, offices, and landscapes. RealEstate10K is useful for tasks such as stereo magnification, light field rendering, and novel view synthesis. However, RealEstate10K also has some challenges, such as the low quality of the videos, the inconsistency of the camera poses, and the lack of depth information\cite{RealEstate10K2018}.
\end{itemize}

Point cloud data sets are collections of points that represent the shape and appearance of a 3D object or scene. They are often obtained from sensors such as lidar, radar, or cameras. Some of the data sets are:

\begin{itemize}
	\item ShapeNet is a large-scale repository of 3D CAD models that covers 55 common object categories and 4 million models. It provides rich annotations such as category labels, part labels, alignments, and correspondences. ShapeNet is useful for tasks such as shape classification, segmentation, retrieval, and completion.Some of the limitations of ShapeNet are that it does not contain realistic textures or materials, it does not capture the variability and diversity of natural scenes, and it does not provide ground truth poses or camera parameters for rendering\cite{ShapeNet}.
	\item KITTI is a data set for autonomous driving that contains 3D point clouds captured by a Velodyne HDL-64E LIDAR sensor, along with RGB images, GPS/IMU data, object annotations, and semantic labels. KITTI is one of the most popular and challenging data sets for 3D object detection and semantic segmentation, as it covers various scenarios, weather conditions, and occlusions. However, KITTI also has some limitations, such as the limited number of frames per sequence (around 200), the fixed sensor configuration, and the lack of dynamic objects\cite{Geiger2012}.
	\item nuScenes is another data set for autonomous driving that contains 3D point clouds captured by a 32-beam LIDAR sensor, along with RGB images, radar data, GPS/IMU data, object annotations, and semantic labels. nuScenes is more comprehensive and diverse than KITTI, as it covers 1000 scenes from six cities in different countries, with varying traffic rules and driving behaviors. nuScenes also provides more temporal information, with 20 seconds of continuous data per scene. However, nuScenes also has some challenges, such as the lower resolution of the point clouds, the higher complexity of the scenes, and the need for sensor fusion\cite{caesar2020nuscenes}.
	\item Matterport3D is a data set for indoor scene understanding that contains 3D point clouds reconstructed from RGB-D images captured by a Matterport camera. The data set also provides surface reconstructions, camera poses, and 2D and 3D semantic segmentations. Matterport3D is a large-scale and high-quality data set that covers 10,800 panoramic views from 194,400 RGB-D images in 90 building types. Matterport3D is useful for tasks such as keypoint matching, view overlap prediction, and scene completion. However, Matterport3D also has some limitations, such as the lack of dynamic objects, the dependence on RGB-D sensors, and the difficulty of obtaining ground truth annotations\cite{ramakrishnan2021hm3d}.
\end{itemize}

Depth map data sets are collections of images and their corresponding depth values, which can be used for various computer vision tasks such as depth estimation, 3D reconstruction, scene understanding, etc.The commonly used depth map data sets are as follows:

\begin{itemize}
	\item Middlebury Stereo is a data set of stereo images with ground truth disparity maps obtained using structured light or a robot arm. It contains several versions of data sets collected from 2001 to 2021, with different scenes, resolutions, and levels of difficulty. The data set is widely used for evaluating stereo matching algorithms and provides online benchmarks and leaderboards. The strengths of this data set are its high accuracy, diversity, and availability. The limitations are its relatively small size, indoor scenes only, and lack of semantic labels\cite{2002Scharstein,2003Scharstein,2007Scharstein,2007Hirschmuller,Scharstein2014HighResolutionSD}.
	\item NYU Depth Data set V2 is a data set of RGB-D images captured by Microsoft Kinect in various indoor scenes. It contains 1449 densely labeled pairs of aligned RGB and depth images, as well as 407024 unlabeled frames. The data set also provides surface normals, 3D point clouds, and semantic labels for each pixel. The data set is widely used for evaluating monocular depth estimation algorithms and provides online tools for data processing and visualization. The strengths of this data set are its large size, rich annotations, and realistic scenes. The limitations are its low resolution, noisy depth values, and indoor scenes only\cite{SilbermanECCV12}.
	\item KITTI also includes depth maps, but its depth maps are limited by sparse and noisy LiDAR depth maps. There is also a lack of real depth maps on the ground for certain scenes, as well as limitations on city Settings\cite{Geiger2012}.
\end{itemize}

Illumination data sets are collections of information about the intensity, distribution, and characteristics of artificial or natural light sources. Some examples of common illumination data sets are:

\begin{itemize}
	\item Multi-PIE is a large-scale data set that contains over 750,000 images of 337 subjects, captured in 15 view angles and 19 illumination conditions. Each subject also performed different facial expressions, such as neutral, smile, surprise, and squint. The data set is useful for studying face recognition, face alignment, face synthesis, and face editing under varying conditions.However, Multi-PIE only contains images of Caucasian subjects, which limits its diversity and generalization\cite{Multi-PIE2008}.
	\item Relightables is a collection of high-quality 3D scans of human subjects under varying lighting conditions. This data set allows for realistic rendering of human performances with any lighting and viewpoint, which can be integrated into any CG scene. Nevertheless, this data set has some drawbacks, such as the low diversity of subjects, poses, and expressions, and the high computational expense of processing the data\cite{Relightables2019}.
\end{itemize}

In conclusion, data sets are essential for 3D controllable image synthesis based on deep learning, as they provide the necessary information for training and evaluating deep generative models. These data sets provide rich annotations and variations for different type of control, such as viewpoint, lighting, poses, point clouds, and depth. However, each data set has its own strengths and weaknesses, and there is still room for improvement and innovation in this field.

\subsection{Evaluation Indicators}

To evaluate the quality and diversity of the synthesized images, several performance indicators are commonly used.  Some of them are:

- Peak signal-to-noise ratio (PSNR)\cite{PSNR}: This measures the similarity between the synthesized image and a reference image in terms of pixel values.  It is defined as the ratio of the maximum possible power of a signal to the power of noise that affects the fidelity of its representation.  A higher PSNR indicates a better image quality.

- Structural similarity index (SSIM)\cite{SSIM}: This measures the similarity between the synthesized image and a reference image in terms of luminance, contrast, and structure.  It is based on the assumption that the human visual system is highly adapted to extract structural information from images.  A higher SSIM indicates a better image quality.

- Learned Perceptual Image Patch Similarity (LPIPS)\cite{LPIPS}: This measures the similarity between the synthesized image and a reference image in terms of deep features. It is defined as the distance between the activations of two image patches for a pre-trained network. A lower LPIPS indicates a better image quality.

- Inception score (IS)\cite{IS}: This measures the quality and diversity of the synthesized images using a pre-trained classifier, such as Inception-v3.  It is based on the idea that good images should have high class diversity (i.e., they can be classified into different categories) and low class ambiguity (i.e., they can be classified with high confidence).  A higher IS indicates a better image synthesis.

- Fréchet inception distance (FID)\cite{FID}: This measures the distance between the feature distributions of the synthesized images and the real images using a pre-trained classifier, such as Inception-v3.  It is based on the idea that good images should have similar feature statistics to real images.  A lower FID indicates a better image synthesis.

- Kernel Inception Distance (KID)\cite{KID}: This measures the squared maximum mean discrepancy between the feature distributions of the synthesized images and the real images using a pre-trained classifier, such as Inception-v3. It is based on the idea that good images should have similar feature statistics to real images. A lower KID indicates a better image synthesis. 

\section{Pose Manipulation}
\label{sec3}

\subsection{Global Pose}

\subsubsection{GAN}

\begin{figure}[b]
	\centering
	\includegraphics[width = 0.4\textwidth]{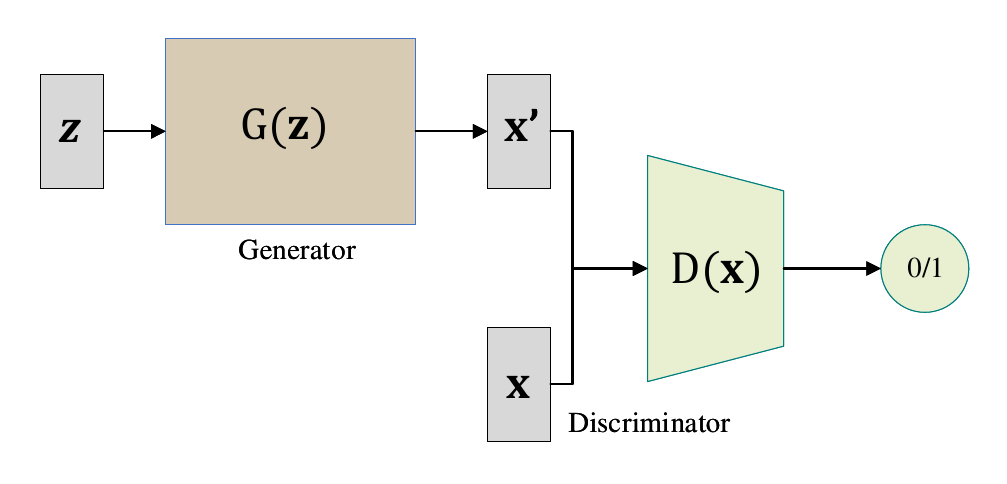}
	\caption{Schematic of GAN, which come from \cite{shi2022deep}}
	\label{fig:5}
\end{figure}

Generative Adversarial Network (GAN)\cite{GAN} can generate realistic and diverse data from a latent space.   GAN consists of two neural networks: a generator and a discriminator.   The generator tries to produce data that can fool the discriminator, while the discriminator tries to distinguish between real and fake data.   The loss function of GAN measures how well the generator and the discriminator perform their tasks.   The loss function is usually composed of two terms: one for the generator($\mathcal{L}_{G}$) and one for the discriminator($\mathcal{L}_{D}$). $\mathcal{L}_{G}$ is based on how often the discriminator classifies the generated data as real, while $\mathcal{L}_{D}$ is based on how often it correctly classifies the real and fake data.   The goal of GAN is to minimize $\mathcal{L}_{G}$ and maximize $\mathcal{L}_{D}$.

$$
\mathcal{L}_{D} =\mathbb{E}_{\mathbf{x} \sim p_{\mathbf{data}}\mathbf{(x)}}[\log (D(\mathbf{x}))]+\mathbb{E}_{\mathbf{z} \sim p_{\mathbf{z}}\mathbf{(z)}}[\log (1-D(G(\mathbf{z})))], 
$$
$$
\mathcal{L}_{G} =-\mathbb{E}_{\mathbf{z} \sim p_{\mathbf{z}}\mathbf{(z)}}[\log (D(G(\mathbf{z})))],
$$
$$
\mathcal{L}_{GAN} =\mathcal{L}_{D}+\mathcal{L}_{G}.
$$

\textbf{a)Crossview image synthesis.}
Viewpoint manipulation refers to the ability to manipulate the perspective or orientation of the objects or scenes in the synthetic images. The earliest view composites were usually only able to composite a specific view, such as a bird's eye view, a frontal view of a person's face, etc. Huang \textit{et~al.} introduced TP-GAN, a method that integrates global structure and local details to generate realistic frontal views of faces \cite{huang2017face}. Similarly, Zhao \textit{et~al.} proposed VariGAN, which combines variational inference and generative adversarial networks for the progressive refinement of synthesized target images \cite{zhao2018multiview}. To address the challenge of generating scenes from different viewpoints and resolutions, Krishna and Ali developed two methods: Crossview Fork (X-Fork) and Crossview Sequential (X-Seq) \cite{regmi2018crossview}. These methods employ semantic segmentation graphs to aid conditional GANs (cGANs) in producing sharper images. Furthermore, Krishna and Ali utilized geometry-guided cGANs for image synthesis, converting ground images to aerial views \cite{Regmi_2019}.Mokhayeri \textit{et~al.} proposed a cross-domain face synthesis approach using a Controllable GAN (C-GAN). This method generates realistic face images under various poses by refining simulated images from a 3D face model through an adversarial game \cite{mokhayeri2019crossdomain}. Zhu \textit{et~al.} developed BridgeGAN, a technique for synthesizing bird's eye view images from single frontal view images. They employed a homography view as an intermediate representation to accomplish this task \cite{zhu2019generative}. Ding \textit{et~al.} addressed the problem of cross-view image synthesis by utilizing generative adversarial networks (GANs) based on deformable convolution and attention mechanisms \cite{ding2020crossview}. Lastly, Ren \textit{et~al.} proposed MLP-Mixer GANs for cross-view image conversion. This method comprises two stages to alleviate severe deformation when generating entirely different views \cite{ren2021cascaded}.

\textbf{b)Free viewpoint image synthesis.}
By adding conditional inputs such as camera pose or camera manifold to the GAN network, it can output images from any viewpoint. Zhu \textit{et~al.} introduced CycleGAN, a method capable of recovering the front face from a single profile postural facial image, even when the source domain does not match the target domain\cite{zhu2020unpaired}. This approach is based on a conditional variational autoencoder and generative adversarial network (cVAE-GAN) framework, which does not require paired data, making it a versatile method for view translation\cite{yin2020novel}.Shen \textit{et~al.} proposed Pairwise-GAN, employing two parallel U-Nets as generators and PatchGAN as a discriminator to synthesize frontal face images\cite{shen2020pairwisegan}. Similarly, Chan \textit{et~al.} presented pi-GAN, a method utilizing periodic implicit Generative Adversarial Networks for high-quality 3D-aware image synthesis\cite{chan2021pigan}. Cai \textit{et~al.} further extended this approach with Pix2NeRF, an unsupervised method leveraging pi-GAN to train on single images without relying on 3D or multi-view supervision\cite{cai2022pix2nerf}.Leimkuhler \textit{et~al.} introduced FreeStyleGAN, which integrates a pre-trained StyleGAN into standard 3D rendering pipelines, enabling stereo rendering or consistent insertion of faces in synthetic 3D environments\cite{Leimk_hler_2021}. Medin \textit{et~al.} proposed MOST GAN, explicitly incorporating physical facial attributes as prior knowledge to achieve realistic portrait image manipulation\cite{medin2021mostgan}.On the other hand, Or-El \textit{et~al.} developed StyleSDF, a novel method generating images based on StyleGAN2 by utilizing Signed Distance Fields (SDFs) to accurately model 3D surfaces, enabling volumetric rendering with consistent results\cite{orel2022stylesdf}. Additionally, Zheng \textit{et~al.} presented SDF-StyleGAN, a deep learning method for generating 3D shapes based on StyleGAN2, employing two new shape discriminators operating on global and local levels to compare real and synthetic SDF values and gradients, significantly enhancing shape geometry and visual quality\cite{zheng2022sdfstylegan}.Moreover, Deng \textit{et~al.} proposed GRAM, a novel approach regulating point sampling and radiance field learning on 2D manifolds, embodied as a set of learned implicit surfaces in the 3D volume, leading to improved synthesis results\cite{deng2022gram}. Xiang \textit{et~al.} built upon this work with GRAM-HD, capable of generating high-resolution images with strict 3D consistency, up to a resolution of 1024x1024\cite{xiang2022gramhd}.In another line of research, Chan \textit{et~al.} developed an efficient framework for generating realistic 3D shapes from 2D images using generative adversarial networks (GANs), comprising a geometry-aware module predicting the 3D shape and its projection parameters from the input image, and a refinement module enhancing shape quality and details\cite{chan2022efficient}. Similarly, Zhao \textit{et~al.} proposed a method for generating high-quality 3D images from 2D inputs using GAN, achieving consistency across different viewpoints and offering rendering with novel lighting effects\cite{zhao2022generative}.Lastly, Alhaija \textit{et~al.} introduced XDGAN, a method for synthesizing realistic and diverse 3D shapes from 2D images, converting 3D shapes into compact 1-channel geometry images and utilizing StyleGAN3 and image-to-image translation networks to generate 3D objects in a 2D space\cite{alhaija2022xdgan}. These advancements in image synthesis techniques have significantly enriched the field of 3D image generation from 2D inputs.

\subsubsection{NeRF}
\begin{figure}[b]
	\centering
	\includegraphics[width = 0.45\textwidth]{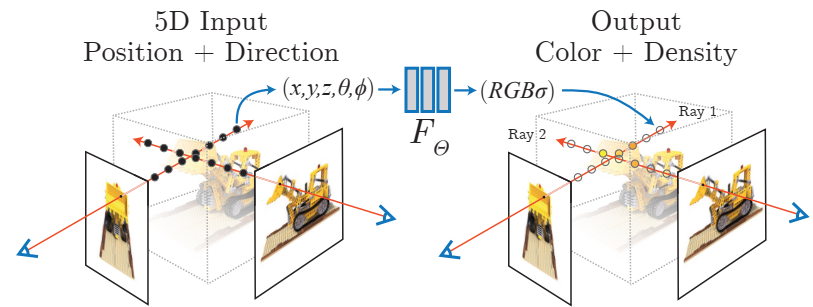}
	\caption{Schematic of NeRF, which come from \cite{NeRF}}
	\label{fig:4}
\end{figure}

Neural Radiance Fields (NeRF)\cite{NeRF} are a novel representation for complex 3D scenes that can be rendered photorealistically from any viewpoint.     NeRF models a scene as a continuous function that maps 5D coordinates (3D location and 2D viewing direction, expressed as ($x$, $y$, $z$, $\theta$, $\varphi$)) to a 4D output (RGB color and opacity).     This function is learned from a set of posed images of the scene using a deep neural network. Before the nerf passes the ($x$, $y$, $z$, $\theta$, $\varphi$) input to the network, it maps the input to a higher dimensional space using high-frequency functions to better fit the data containing high-frequency variations.  The high-frequency coding function is:
$$
\left(\sin \left(2^{0} \pi p\right),\cos \left(2^{0} \pi p\right),\ldots,\sin \left(2^{L-1} \pi p\right),\cos \left(2^{L-1} \pi p\right)\right)
$$
Where $p$ is the input ($x$, $y$, $z$, $\theta$, $\varphi$).

\begin{table*}[t]
	\begin{center}
		\renewcommand{\arraystretch}{1.2}
		\caption{Enhancements to NeRF}\label{tab1}
    \resizebox{2\columnwidth}{!}{
		\begin{tabular}{|c|c|c|c|c|}
			\hline
														 & Method                                 & Publication & Image resolution                          & Dataset                                                                                                          \\ \hline
			\multirow{5}{*}{no camera pose}              & NeRF--\cite{wang2022nerf}              & arXiv2022   & 756x1008/1080x1920/520x780                & \cite{Mildenhall2019LLFF}/\cite{RealEstate10K2018}/\cite{wang2022nerf}                                           \\ \cline{2-5} 
														 & GNeRF\cite{meng2021gnerf}              & ICCV2021    & 400x400/500x400                           & \cite{NeRF}/\cite{DTU}                                                                                           \\ \cline{2-5} 
														 & SCNeRF\cite{SCNeRF2021}                & ICCV2021    &                                           & \cite{Mildenhall2019LLFF}/\cite{Tanks_and_Temples}                                                               \\ \cline{2-5} 
														 & NoPe-NeRF\cite{bian2022nopenerf}       & CVPR2023    &                                           & \cite{DTU}/\cite{Mildenhall2019LLFF}/\cite{Replica}                                                              \\ \cline{2-5} 
														 & SPARF\cite{sparf2023}                  & CVPR2023    & 960x540/648x484                           & \cite{Tanks_and_Temples}/\cite{dai2017scannet}                                                                   \\ \hline
			\multirow{5}{*}{sparse data}                 & NeRS\cite{zhang2021ners}               & NIPS2021    &                                           & \cite{zhang2021ners}                                                                                             \\ \cline{2-5} 
														 & MixNeRF\cite{seo2023mixnerf}           & CVPR2023    &                                           & \cite{DTU}/\cite{Mildenhall2019LLFF}/\cite{NeRF}                                                                 \\ \cline{2-5} 
														 & SceneRF\cite{cao2023scenerf}           & arXiv2023   & 1220x370                                  & \cite{behley2019iccv}                                                                                            \\ \cline{2-5} 
														 & GM-NeRF\cite{chen2023gmnerf}           & CVPR2023    & 224x224                                   & \cite{tao2021function4d}/\cite{Multi-garment}/\cite{Genebody}/\cite{ZJUMocap}                                    \\ \cline{2-5} 
														 & SPARF\cite{sparf2023}                  & CVPR2023    & 960x540/648x484                           & \cite{Tanks_and_Temples}/\cite{dai2017scannet}                                                                   \\ \hline
			\multirow{4}{*}{noisy data}                  & RawNeRF\cite{mildenhall2021nerf}       & CVPR2022    &                                           & \cite{mildenhall2021nerf}                                                                                        \\ \cline{2-5} 
														 & Deblur-NeRF\cite{ma2022deblurnerf}     & CVPR2022    &                                           & \cite{ma2022deblurnerf}                                                                                          \\ \cline{2-5} 
														 & HDR-NeRF\cite{huang2022hdr}            & CVPR2022    & 400x400/804x534                           & \cite{huang2022hdr}                                                                                              \\ \cline{2-5} 
														 & NAN\cite{pearl2022nan}                 & CVPR2022    &                                           & \cite{pearl2022nan}                                                                                              \\ \hline
			\multirow{5}{*}{large-scale image synthesis} & Mip-NeRF 360\cite{barron2022mipnerf}   & CVPR2022    &                                           & \cite{Tanks_and_Temples}                                                                                         \\ \cline{2-5} 
														 & BungeeNeRF\cite{xiangli2022bungeenerf} & ECCV2022    &                                           & \cite{googleearthstudio}                                                                                         \\ \cline{2-5} 
														 & Block-NeRF\cite{tancik2022blocknerf}   & arXiv2022   &                                           & \cite{tancik2022blocknerf}                                                                                       \\ \cline{2-5} 
														 & GridNeRF\cite{xu2023gridguided}        & CVPR2023    &                                           & \cite{Rubble}/\cite{xu2023gridguided}                                                                            \\ \cline{2-5} 
														 & EgoNeRF\cite{choi2023balanced}         & CVPR2023    &                                           & \cite{choi2023balanced}                                                                                          \\ \hline
			\multirow{6}{*}{image synthesis speed}       & PlenOctrees\cite{yu2021plenoctrees}    & ICCV2021    & 800x800/1920x1080                         & \cite{NeRF}/\cite{Tanks_and_Temples}                                                                             \\ \cline{2-5} 
														 & DirectVoxGO\cite{sun2022direct}        & CVPR2022    & 800x800/800x800/768x576/1920x1080/512x512 & \cite{NeRF}/\cite{liu2020neural}/\cite{yao2020blendedmvs}/\cite{Tanks_and_Temples}/\cite{sitzmann2019deepvoxels} \\ \cline{2-5} 
														 & R2L\cite{wang2022r2l}                  & ECCV2022    &                                           & \cite{NeRF}/\cite{neff2021donerf}                                                                                \\ \cline{2-5} 
														 & SqueezeNeRF\cite{Wadhwani_2022}        & CVPR2022    &                                           & \cite{NeRF}/\cite{Mildenhall2019LLFF}                                                                            \\ \cline{2-5} 
														 & MobileNeRF\cite{chen2023mobilenerf}    & CVPR2023    & 800x800/1008x756/1256x828                 & \cite{NeRF}/\cite{Mildenhall2019LLFF}/\cite{barron2022mipnerf}                                                   \\ \cline{2-5} 
														 & L2G-NeRF\cite{chen2023localtoglobal}   & CVPR2023    &                                           & \cite{Mildenhall2019LLFF}                                                                                        \\ \hline
			\end{tabular}}
	\end{center}
\end{table*}

Zhang \textit{et~al.} introduced NeRF++ as a framework that enhances NeRF (Neural Radiance Fields) through adaptive sampling, hierarchical volume rendering, and multi-scale feature encoding techniques \cite{zhang2020nerf}. This approach enables high-quality rendering for both static and dynamic scenes while improving efficiency and robustness. Rebain \textit{et~al.} proposed a method to enhance the efficiency and quality of neural rendering by employing spatial decomposition \cite{rebain2020derf}.Park \textit{et~al.} developed a novel technique for capturing and rendering high-quality 3D selfies using a single RGB camera. Their method utilizes a deformable neural radiance field (NeRF) model capable of representing both the geometry and appearance of dynamic scenes \cite{park2021nerfies}. Li \textit{et~al.} introduced MINE, a method for novel view synthesis and depth estimation from a single image. This approach generalizes Multiplane Images (MPI) with continuous depth using Neural Radiance Fields (NeRF) \cite{li2021mine}.Park \textit{et~al.} proposed HyperNeRF, a method for representing and rendering complex 3D scenes with varying topology using neural radiance fields (NeRFs). Unlike previous NeRF-based approaches that rely on a fixed 3D coordinate system, HyperNeRF employs a higher-dimensional continuous embedding space to capture arbitrary scene changes \cite{park2021hypernerf}. Chen \textit{et~al.} presented Aug-NeRF, a novel method for training neural radiance fields (NeRFs) with physically-grounded augmentations at different levels: scene, camera, and pixel \cite{chen2022augnerf}.Kaneko proposed AR-NeRF, a method for learning 3D representations of natural images without supervision. The approach utilizes a neural radiance field (NeRF) model to render images with various viewpoints and aperture sizes, capturing both depth and defocus effects \cite{kaneko2022arnerf}. Li \textit{et~al.} introduced SymmNeRF, a framework that utilizes neural radiance fields (NeRFs) to synthesize novel views of objects from a single image. This method leverages symmetry priors to recover fine appearance details, particularly in self-occluded areas \cite{li2023symmnerf}.Zhou \textit{et~al.} proposed NeRFLiX, a novel framework for improving the quality of novel view synthesis using neural radiance fields (NeRF). This approach addresses rendering artifacts such as noise and blur by employing an inter-viewpoint aggregation framework that fuses high-quality training images to generate more realistic synthetic views \cite{zhou2023nerflix}.

Besides, a number of researchers have proposed enhancements to the original NeRF model, addressing its limitations in scenarios such as no camera pose, sparse data, noisy data, large-scale image synthesis, and image synthesis speed.See Table \ref{tab1}.

\subsubsection{Diffusion model}

\begin{figure}[b]
	\centering
	\includegraphics[width = 0.45\textwidth]{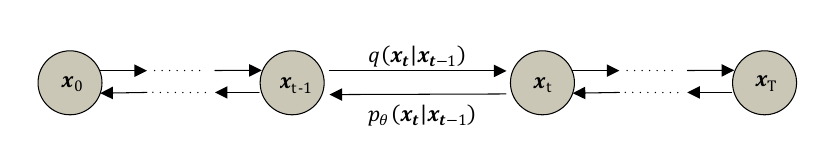}
	\caption{Schematic of diffusion model, which come from \cite{shi2022deep}}
	\label{fig:7}
\end{figure}

One of the most widely used models in deep learning is the diffusion model, which is a generative model that can produce realistic and diverse images from random noise.      The diffusion model is based on the idea of reversing the process of adding Gaussian noise to an image until it becomes completely corrupted.   The diffusion process starts from a data sample and gradually adds noise until it reaches a predefined noise level,  i.e. $q\left(\mathbf{x}_{t} \mid \mathbf{x}_{t-1}\right)$ in the Fig. \ref{fig:7}.    The generative model then learns to reverse this process by denoising the samples at each step  i.e. $p_{\mathbf{\theta }}\left(\mathbf{x}_{t-1} \mid \mathbf{x}_{t}\right)$ in the Fig. \ref{fig:7}.
$$
q\left(\mathbf{x}_{t} \mid \mathbf{x}_{t-1}\right)=\mathcal{N}\left(\mathbf{x}_{t} ; \sqrt{1-\beta_{t}} \mathbf{x}_{t-1}, \beta_{t} \mathbf{I}\right)
$$

Sbrolli \textit{et~al.} introduced IC3D, a novel approach addressing various challenges in shape generation. This method is capable of reconstructing a 3D shape from a single view, synthesizing a 3D shape from multiple views, and completing a 3D shape from partial inputs \cite{sbrolli2023ic3d}. Another significant contribution in this area is the work by Gu \textit{et~al.}, who developed Control3Diff, a generative model with 3D-awareness and controllability. By combining diffusion models and 3D GANs, Control3Diff can synthesize diverse and realistic images without relying on 3D ground truth data, and can be trained solely on single-view image datasets \cite{gu2023learning}.Additionally, Anciukevicius \textit{et~al.} proposed RenderDiffusion, an innovative diffusion model for 3D generation and inference. Remarkably, this model can be trained using only monocular 2D supervision and incorporates an intermediate three-dimensional representation of the scene during each denoising step, effectively integrating a robust inductive structure into the diffusion process \cite{anciukevicius2023renderdiffusion}.Xiang \textit{et~al.} presented a novel method for generating 3D-aware images using 2D diffusion models. Their approach involves a sequential process of generating multiview 2D images from different perspectives, ultimately achieving the synthesis of 3D-aware images \cite{xiang20233daware}.Furthermore, Liu \textit{et~al.} proposed a framework for changing the camera viewpoint of an object using only a single RGB image. Leveraging the geometric priors learned by large-scale diffusion models about natural images, their framework employs a synthetic dataset to learn the controls for adjusting the relative camera viewpoint \cite{liu2023zero1to3}.Lastly, Chan \textit{et~al.} developed a method for generating diverse and realistic novel views of a scene based on a single input image. Their approach utilizes a diffusion-based model that incorporates 3D geometry priors through a latent feature volume. This feature volume captures the distribution of potential scene representations and enables the rendering of view-consistent images \cite{chan2023generative}.

\begin{figure}[b]
	\centering
	\includegraphics[width = 0.45\textwidth]{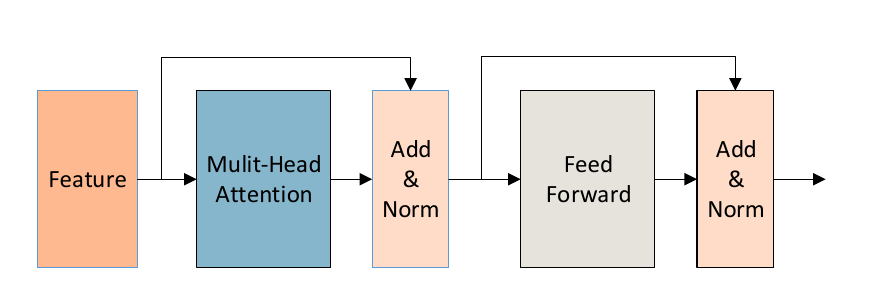}
	\caption{Schematic of Transformer, which come from \cite{Vaswani2017}}
	\label{fig:6}
\end{figure}
\subsubsection{Transformer}
Transformers are a type of neural network architecture that have been widely used in natural language processing.    They are based on the idea of self-attention, which allows the network to learn the relationships between different parts of the input and output sequences.  Transformers is introduced into the field of computer vision in the paper ViT\cite{dosovitskiy2020vit}.  Its core is the Attention section in Fig. \ref{fig:6}, and its formula is as follows:
$$
Attention(Q,K,V)=softmax(\frac{QK^T}{\sqrt{d_k}})V
$$

Leveraging the Transformer architecture for vision applications, several studies have explored its potential for synthesizing 3D views. Nguyen-Ha and colleagues presented a pioneering approach to synthesizing new views of a scene using a given set of input views. Their method employs a transformer-based architecture that effectively captures the long-range dependencies among the input views. By using a sequential process, the method generates high-quality novel views. This research contribution is documented in \cite{nguyenha2020sequential}. Similarly, Yang and colleagues proposed an innovative method for generating viewpoint-invariant 3D shapes from a single image. Their approach is based on disentangling learning and parametric NURBS surface generation. The method employs an encoder-decoder network augmented with a disentangled transformer module. This configuration enables the independent learning of shape semantics and camera viewpoints. The output of this comprehensive network includes the geometric parameters of the NURBS surface representing the 3D shape, as well as the camera-viewpoint parameters involving rotation, translation, and scaling. Further details of this method can be found in \cite{jinglun2021}. Additionally, Kulhánek and colleagues proposed ViewFormer, an impressive neural rendering method that does not rely on NeRF and instead capitalizes on the power of transformers. ViewFormer is designed to learn a latent representation of a scene using only a few images, and this learned representation enables the synthesis of novel views. Notably, ViewFormer can handle complex scenes with varying illumination and geometry without requiring any 3D information or ray marching. The specific approach and findings of ViewFormer are detailed in \cite{kulh2022viewformer}.

\subsubsection{Hybrid NeRF}
\textbf{a)GAN-based NeRF.}
Neural Radiance Fields (NeRF) are a novel method for rendering images from arbitrary viewpoints, but they suffer from high computational cost due to their pixel-wise optimization. Generative Adversarial Networks (GANs) can synthesize realistic images in a single forward pass, but they may not preserve the view consistency across different viewpoints. Hence, there is a growing interest in exploring the integration of NeRF and GAN for efficient and consistent image synthesis.Meng \textit{et~al.} presented the GNeRF framework, which combines GANs and NeRF reconstruction to generate scenes with unknown or random camera poses \cite{meng2021gnerf}. Similarly, Zhou \textit{et~al.} introduced CIPS-3D, a generative model that utilizes style transfer, shallow NeRF networks, and deep INR networks to represent 3D scenes and provide precise control over camera poses \cite{zhou2021cips3d}.Another approach by Xu \textit{et~al.} is GRAF, a generative model for radiance fields that enables high-resolution image synthesis while being aware of the 3D shape. GRAF disentangles camera and scene properties from unposed 2D images, allowing for the synthesis of novel views and modifications to shape and appearance \cite{xu2021generative}. Lan \textit{et~al.} proposed a self-supervised geometry-aware encoder for style-based 3D GAN inversion. Their encoder recovers the latent code of a given 3D shape and enables manipulation of its style and geometry attributes \cite{lan2022selfsupervised}.Li \textit{et~al.} developed a two-step approach for 3D-aware multi-class image-to-image translation using NeRFs. They trained a multi-class 3D-aware GAN with a conditional architecture and innovative training strategy. Based on this GAN, they constructed a 3D-aware image-to-image translation system \cite{li20233daware}. Shahbazi \textit{et~al.} focused on knowledge distillation, proposing a method to transfer the knowledge of a GAN trained on NeRF representation to a convolutional neural network (CNN). This enables efficient 3D-aware image synthesis \cite{shahbazi2023nerfgan}.Kania \textit{et~al.} introduced a generative model for 3D objects based on NeRFs, which are rendered into 2D novel views using a hypernetwork. The model is trained adversarially with a 2D discriminator \cite{kania2023}. Lastly, Bhattarai \textit{et~al.} proposed TriPlaneNet, an encoder specifically designed for EG3D inversion. The task of EG3D inversion involves reconstructing 3D shapes from 2D edge images \cite{bhattarai2023triplanenet}. 

\textbf{b)Diffusion model-based NeRF.}
Likewise, the diffusion model alone fails to produce images that are consistent across different viewpoints. Therefore, many researchers integrate it with NeRF to synthesize high-quality and view-consistent images.Muller \textit{et~al.} proposed DiffRF, which directly generates volumetric radiance fields from a set of posed images by using a 3D denoising model and a rendering loss \cite{muller2023diffrf}. Similarly, Xu \textit{et~al.} proposed NeuralLift-360, a framework that generates a 3D object with 360° views from a single 2D photo using a depth-aware NeRF and a denoising diffusion model \cite{xu2023neurallift360}. Chan \textit{et~al.} proposed a 3D-aware image synthesis framework using NeRF and diffusion models, which jointly optimizes a NeRF auto-decoder and a latent diffusion model to enable simultaneous 3D reconstruction and prior learning from multi-view images of diverse objects \cite{chen2023singlestage}. Lastly, Gu \textit{et~al.} proposed NerfDiff, a method for generating realistic and 3D-consistent novel views from a single input image. This method distills the knowledge of the conditional diffusion model (CDM) into the NeRF by synthesizing and refining a set of virtual views at test time, using a NeRF-guided distillation algorithm \cite{gu2023nerfdiff}. These approaches demonstrate the potential of using NeRF and diffusion models for 3D scene synthesis, and further research in this area is expected to yield even more exciting results.

\textbf{c)Transformer-based NeRF.}
Building on the previous work of integrating generative adversarial networks (GANs) and neural radiance fields (NeRFs), some researchers have explored the possibility of using Transformer models and NeRFs to generate 3D images that are consistent across different viewpoints. Wang \textit{et~al.} proposed a method that can handle complex scenes with dynamic objects and occlusions, and can generalize to unseen scenes without fine-tuning. The key idea is to use a transformer to learn a global latent representation of the scene, which is then used to condition a NeRF model that renders novel views \cite{wang2022generalizable}. Similarly, Lin \textit{et~al.} proposed a method for novel view synthesis from a single unposed image using NeRF and vision transformer (ViT). The method leverages both global and local image features to form a 3D representation of the scene, which is then used to render novel views by a multi-layer perceptron (MLP) network \cite{lin2022vision}. Finally, Liu \textit{et~al.} proposed a method for visual localization using a conditional NeRF model. The method can estimate the 6-DoF pose of a query image given a sparse set of reference images and their poses \cite{liu2023nerfloc}. These methods demonstrate the potential of NeRFs and transformers in addressing challenging problems in computer vision.

\subsection{Local Pose}
\subsubsection{GAN}Liao \textit{et~al.} proposed a novel framework consisting of two components for learning generative models that can achieve this goal. The first component is a 3D generator that learns to reconstruct the 3D shape and appearance of an object from a single image, while the second component is a 2D generator that learns to render the 3D object into a 2D image. This framework can generate high-quality images with controllable factors such as pose, shape, and appearance \cite{Liao2020CVPR}.Nguyen-Phuoc \textit{et~al.} proposed BlockGAN, a novel image generative model that can create realistic images of scenes composed of multiple objects. BlockGAN learns to generate 3D features for each object and combine them into a 3D scene representation. The model then renders the 3D scene into a 2D image, taking into account the occlusion and interaction between objects, such as shadows and lighting. BlockGAN can manipulate the pose and identity of each object independently while preserving image quality \cite{BlockGAN2020}.Pan \textit{et~al.} proposed a novel framework that can reconstruct 3D shapes from 2D image GANs without any supervision or prior knowledge. The method can generate realistic and diverse 3D shapes for various object categories, and the reconstructed shapes are consistent with the 2D images generated by the GANs. The recovered 3D shapes allow high-quality image editing such as relighting and object rotation \cite{pan2020gan2shape}.Tewari \textit{et~al.} proposed a novel 3D generative model that can learn to separate the geometry and appearance factors of objects from a dataset of monocular images. The model uses a non-rigid deformable scene formulation, where each object instance is represented by a deformed canonical 3D volume. The model can also compute dense correspondences between images and embed real images into its latent space, enabling editing of real images \cite{tewari2022disentangled3d}.

\subsubsection{NeRF}Niemeyer and Geiger introduced GIRAFFE, a deep generative model that can synthesize realistic and controllable images of 3D scenes. The model represents scenes as compositional neural feature fields that encode the shape and appearance of individual objects as well as the background. The model can disentangle these factors from unstructured and unposed image collections without any additional supervision. With GIRAFFE, individual objects in the scene can be manipulated by translating, rotating, or changing their appearance, as well as changing the camera pose\cite{Niemeyer2020GIRAFFE}.Yang \textit{et~al.} proposed a neural scene rendering system called OC-Nerf that learns an object-compositional neural radiance field for editable scene rendering. OC-Nerf consists of a scene branch and an object branch, which encode the scene and object geometry and appearance, respectively. The object branch is conditioned on learnable object activation codes that enable object-level editing such as moving, adding, or rotating objects\cite{yang2021objectnerf}.Kobayashi \textit{et~al.} proposed a method to enable semantic editing of 3D scenes represented by neural radiance fields (NeRFs). The authors introduced distilled feature fields (DFFs), which are 3D feature descriptors learned by transferring the knowledge of pre-trained 2D image feature extractors such as CLIP-LSeg or DINO. DFFs allow users to query and select specific regions or objects in the 3D space using text, image patches, or point-and-click inputs. The selected regions can then be edited in various ways, such as rotation, translation, scaling, warping, colorization, or deletion\cite{kobayashi2022distilledfeaturefields}.Zhang \textit{et~al.} introduced Nerflets, a new approach to represent 3D scenes from 2D images using local radiance fields. Unlike prior approaches that rely on global implicit functions, Nerflets partition the scene into a collection of local coordinate frames that encode the structure and appearance of the scene. This enables efficient rendering and editing of complex scenes with high fidelity and detail. Nerflets can manipulate the object's orientation, position, and size, among other operations\cite{zhang2022nerfusion}.Finally, Zheng \textit{et~al.} proposed EditableNeRF, a method that allows users to edit dynamic scenes modeled by neural radiance fields (NeRF) with key points. The method can handle topological changes and generate novel views from a single camera input. The key points are detected and optimized automatically by the network, and users can drag them to modify the scene. These approaches provide various means for 3D scene synthesis and editing, including manipulating objects, changing camera pose, selecting and editing specific regions or objects, and handling topological changes\cite{zheng2023editablenerf}.

\section{Structure Manipulation}
\label{sec4}
\subsection{Global Structure}
\subsubsection{Editting point cloud}A depth map is a representation of the distance between a scene and a reference point, such as a camera.  It can be used to create realistic effects such as depth of field, occlusion, and parallax\cite{zhang2021}. Chen \textit{et~al.} proposed a novel method called SPIDR for representing and manipulating 3D objects using neural point fields (NPFs) and signed distance functions (SDFs) \cite{bib15}. The method combines explicit point cloud and implicit neural representations to enable high-quality mesh and surface reconstruction for object deformation and lighting estimation. With the trained SPIDR model, various geometric edits can be applied to the point cloud representation, which can be used for image editing.Zhang \textit{et~al.} introduced a new method for rendering point clouds with frequency modulation, which enables easy editing of shape and appearance \cite{zhang2023frequency}. The method converts point clouds into a set of frequency-modulated signals that can be rendered efficiently using Fourier analysis. The signals can also be manipulated in the frequency domain to achieve various editing effects, such as deformation, smoothing, sharpening, and color adjustment.Chen \textit{et~al.} also proposed NeuralEditor, a novel method for editing neural radiance fields (NeRFs) for shape editing tasks \cite{neuraleditor}. The method uses point clouds as the underlying structure to construct NeRFs and renders them with a new scheme based on K-D tree-guided voxels. NeuralEditor can perform shape deformation and scene morphing by mapping points between point clouds.

\subsubsection{Editting depth map}Chen \textit{et~al.} introduced the Visual Object Networks (VON) framework, which enables the disentangled learning of 3D object representations from 2D images. This framework comprises three modules, namely a shape generator, an appearance generator, and a rendering network. By manipulating the generators, VON can perform a range of tasks, including shape manipulation, appearance transfer, and novel view synthesis \cite{VON}.Mirzaei \textit{et~al.} proposed a reference-guided controllable inpainting method for neural radiance fields (NeRFs), which allows for the synthesis of novel views of a scene with missing regions. The method employs a reference image to guide the inpainting process, and a user interface that enables the user to adjust the degree of blending between the reference and the original NeRF \cite{mirzaei2023referenceguided}.Yin \textit{et~al.} introduced OR-NeRF, a novel pipeline that can remove objects from 3D scenes using point or text prompts on a single view. This pipeline leverages a points projection strategy, a 2D segmentation model, 2D inpainting methods, and depth supervision and perceptual loss to achieve better editing quality and efficiency than previous works \cite{yin2023ornerf}.
Chen \textit{et~al.} proposed a visual comfort aware-reinforcement learning (VCARL) method for depth adjustment of stereoscopic 3D images. This method aims to improve the visual quality and comfort of 3D images by learning a depth adjustment policy from human feedback \cite{Kim_Park_Lee_Kim_Ro_2021}. These advancements offer various means of manipulating objects, adjusting depth, and generating novel views, ultimately enhancing the quality and realism of 3D scene synthesis and editing.

\subsection{Local Structure}
\subsubsection{GAN}In recent years, there have been significant advancements in the field of 3D scene inpainting and editing using generative adversarial networks (GANs). Jheng \textit{et~al.} proposed a dual-stream GAN for free-form 3D scene inpainting. The network comprises two streams, namely a depth stream and a color stream, which are jointly trained to inpaint the missing regions of a 3D scene. The depth stream predicts the depth map of the scene, while the color stream synthesizes the color image. This approach enables the removal of objects using existing 3D editing tools \cite{Jheng2022Freeform3S}.Another recent development in GAN training is the introduction of LinkGAN, a regularizer proposed by Zhu \textit{et~al.} that links some latent axes to image regions or semantic categories. By resampling partial latent codes, this approach enables local control of GAN generation \cite{zhu2023linkgan}.Wang \textit{et~al.} proposed a novel method for synthesizing realistic images of indoor scenes with explicit camera pose control and object-level editing capabilities. This method builds on BlobGAN, a 2D GAN that disentangles individual objects in the scene using 2D blobs as latent codes. To extend this approach to 3D scenes, the authors introduced 3D blobs, which capture the 3D nature of objects and allow for flexible manipulation of their location and appearance \cite{wang2023blobgan3d}. These recent advancements in GAN-based 3D scene inpainting and editing have the potential to significantly improve the quality and realism of synthesized scenes.

\subsubsection{NeRF}Liu \textit{et~al.} \cite{liu2020neural} introduced Neural Sparse Voxel Fields (NSVF), which combines neural implicit functions with sparse voxel octrees to enable high-quality novel view synthesis from a sparse set of input images, without requiring explicit geometry reconstruction or meshing. Gu \textit{et~al.} \cite{gu2021stylenerf} introduced StyleNeRF, a method that enables camera pose manipulation for synthesizing high-resolution images with strong multi-view coherence and photorealism. Wang \textit{et~al.} \cite{wang2021clip} introduced CLIP-NeRF, a method for manipulating 3D objects represented by neural radiance fields (NeRF) using text or image inputs. Kania \textit{et~al.} \cite{kania2022conerf} proposed a novel method for manipulating neural 3D representations of scenes beyond novel view rendering by allowing the user to specify which part of the scene they want to control with mask annotations in the training images. Lazova \textit{et~al.} \cite{lazova2022controlnerf} proposed a novel method for performing flexible, 3D-aware image content manipulation while enabling high-quality novel view synthesis by combining scene-specific feature volumes with a general neural rendering network. Yuan \textit{et~al.} \cite{yuan2022nerfediting} proposed a method for user-controlled shape deformation of scenes represented by implicit neural rendering, especially Neural Radiance Field (NeRF). Sun \textit{et~al.} \cite{sun2022nerfeditor} proposed NeRFEditor, a learning framework for 3D scene editing that uses a pre-trained StyleGAN model and a NeRF model to generate stylized images from a 360-degree video input. Wang \textit{et~al.} \cite{wang2022generative} proposed a novel method for image synthesis of topology-varying objects using generative deformable radiance fields (GDRFs). Tertikas \textit{et~al.} \cite{Tertikas2023CVPR} proposed PartNeRF, a novel part-aware generative model for editable 3D shape synthesis that does not require any explicit 3D supervision. Bao \textit{et~al.} \cite{bao2023sine} proposed SINE, a novel approach for editing a neural radiance field (NeRF) with a single image or text prompts. Cohen-Bar \textit{et~al.} \cite{cohenbar2023setthescene} proposed a novel framework for synthesizing and manipulating 3D scenes from text prompts and object proxies. Finally, Mirzaei \textit{et~al.} \cite{mirzaei2023spinnerf} proposed a novel method for reconstructing 3D scenes from multiview images by leveraging neural radiance fields (NeRF) to model the geometry and appearance of the scene, and introducing a segmentation network and a perceptual inpainting network to handle occlusions and missing regions. These methods represent significant progress towards the goal of enabling high-quality, user-driven 3D scene synthesis and editing.

\subsubsection{Diffusion model}Avrahami \textit{et~al.} \cite{Avrahami_2022_CVPR} introduced a method for local image editing based on natural language descriptions and region-of-interest masks. The method uses a pretrained language-image model (CLIP) and a denoising diffusion probabilistic model (DDPM) to produce realistic outcomes that conform to the text input. It can perform various editing tasks, such as object addition, removal, replacement, or modification, background replacement, and image extrapolation.Nichol \textit{et~al.} \cite{nichol2022glide} proposed GLIDE, a diffusion-based model for text-conditional image synthesis and editing. This method uses a guidance technique to trade off diversity for fidelity and produces photorealistic images that match the text prompts.Couairon \textit{et~al.} \cite{couairon2022diffedit} proposed DiffEdit, a method that uses text-conditioned diffusion models to edit images based on text queries. It can automatically generate a mask that highlights the regions of the image that need to be changed according to the text query. It also uses latent inference to preserve the content in those regions. DiffEdit can produce realistic and diverse semantic image edits for various text prompts and image sources.Sella \textit{et~al.} \cite{sella2023voxe} proposed Vox-E, a novel framework that uses latent diffusion models to edit 3D objects based on text prompts. It takes 2D images of a 3D object as input and learns a voxel grid representation of it. It then optimizes a score distillation loss to align the voxel grid with the text prompt while regularizing it in 3D space to preserve the global structure of the original object. Vox-E can create diverse and realistic edits.Haque \textit{et~al.} \cite{instructnerf2023} proposed a novel method for editing 3D scenes with natural language instructions. The method leverages a neural radiance field (NeRF) representation of the scene and a transformer-based model that can parse the instructions and modify the NeRF accordingly. The method can perform various editing tasks, such as changing the color, shape, position, and orientation of objects, as well as adding and removing objects, with high fidelity and realism.Lin \textit{et~al.} \cite{lin2023componerf} proposed CompoNeRF, a novel method for text-guided multi-object compositional NeRF with editable 3D scene layout. CompoNeRF can synthesize photorealistic images of complex scenes from natural language descriptions and user-specified camera poses. It can also edit the 3D layout of the scene by manipulating the objects' positions, orientations, and scales. These methods have shown promising results in advancing the field of image and 3D object editing using natural language descriptions, and they have the potential to be applied in various applications.

\section{Illumination Manipulation}
\label{sec5}
controlable image generation refers to the use of technology to generate images and to constrain and adjust the generation process so that the generated images meet specific requirements.By guiding external conditions or manipulating and adjusting the code, it is possible to trim a certain area or attribute of the image while leaving other areas or attributes unchanged.To solve the low-level image generation problem, we analyze the image generation for different conditions, lighting being one of them, and summarize the algorithms for each solution under different lighting conditions.

\textbf{Inverse rendering.}Currently,neural rendering is applied to scene restruction.One approach is to capture photometric appearance variations in in-the-wild data, decomposing the scene into image-dependent shared components\cite{bib12}.

Another very important type of rendering is inverse rendering.The inverse rendering of objects under completely unknown capture conditions is a fundamental challenge in computer vision and graphics.This challenge is especially acute when the input image is captured in a complex and changing environment.Without using the nerf method,Mark Boss \textit{et~al.} proposed a join optimization framework to estimate the shape,BRDF,per-image camera pose and illumination\cite{bib13}.

Changwoon Choi \textit{et~al.} proposed IBL-NeRF also based on rendering .This method's inverse rendering extends the original NeRF formulation to capture the spatial variation of lighting within the scene volume, in addition to surface properties. Specifically, the scenes of diverse materials are decomposed into intrinsic components for image-based ren dering, namely, albedo, roughness, surface normal, irradiance, and prefiltered radiance.All of the components are inferred as neural images from MLP, and model large-scale general scenes\cite{bib14}.

However, NeRF-based methods encode shape,reflectance and illumination implicitly and this makes it challenging for users to manipulate these properties in the rendered images explicitly.So a new hybrid SDF-based 3D neural representation is generated, capable of rendering scene deformations and lighting more accurately.This neural representation also adds a new SDF regularization.The disadvantage of this approach is that it sacrifices rendering quality. In reverse rendering, high render quality is often at odds with accurate lighting decomposition, as shadows and lighting can easily be misinterpreted as textures. Therefore, rendering quality still requires a concerted effort of surface reconstruction and reverse rendering\cite{bib15}.
While dynamic Neural Radiation Field (NeRF) is a powerful algorithm capable of rendering photo-realistic novel view images from a monocular RGB video of a dynamic scene. But dynamic NeRF does not model the change of the reflected color during the warping.This is one of its drawbacks.To address this problem in rendering, Zhiwen Yan \textit{et~al.} allowed specularly reflective surfaces of different poses to maintain different reflective colors when mapped to the common canonical space by reformulating the neural radiation field function as conditional on the position and orientation of the surface in the observation space.This method more accurately reconstructs and renders dynamic specular scenes\cite{bib16}.

The inverse rendering objective function of this method is as follows:

$$
\mathcal{L}=\mathcal{L}_{\text {render }}+\mathcal{L}_{\text {pref }}+\mathcal{L}_{\text {prior }}+\lambda_{I, \text { reg }} \mathcal{L}_{I, \text { reg }}
$$

$\mathcal{L}_{\text {render }}$ and $\mathcal{L}_{\text {pref }}$ are rendering losses to match the rendered images with the input images. 

Next, we will explain each of these parameters.

$$
\mathcal{L}_{\text {render }}=\left\|L_o(r)-\hat{L}_o(r)\right\|_2^2,
$$

This is for each pixel of the camera light. $r$ represents a single piexl.where $L_o$ is our nated radiance and $\hat{L}_o$ is ground truth radiance.

$$
\mathcal{L}_{\text {pref}}=\sum_j\left\|L_{\text {pref }}^j(r)-L_{\mathrm{G}}^j(r)\right\|_2^2 .
$$
This is the rendering loss of pre-filtered radiation.$L_{\text {pref }}^j(r)$ is inferred prefiltered radiance of $j^(th)$ level and $L_{\mathrm{G}}^j(r)$ is the radiance convolved with jth level Gaussian convolution,where $L_{\mathrm{G}}^0$ = L.

$$
\mathcal{L}_{\text {prior }}=\|a(r)-\hat{a}(r)\|_2^2 .
$$

The equation encourages our inferred albedo $a$ to match the pseudo albedo.

$$
\mathcal{L}_{I, \text { reg }}=\|I(r)-\mathbb{E}[\hat{I}]\|_2^2,
$$

This is the irradiance regularization loss,where $\mathbb{E}[\hat{I}]$ is the mean of irradiance (shading) values in training set images.

\subsection{Global Illumination}

The absence of ideal light and the fact that the studied objects are in an unfavorable environment such as deflection, movement, darkness, and high interference can lead to under-illuminated, single irradiated light source, and complex illumination of the acquired images, all of which can degrade the performance of the final image generation.Next, we will review the various ways to deal with these aspects.

\subsubsection{Brightness level}The use of illumination normalization generative adversarial network - IN-GAN can be well generalized to images with less illumination variations.The method combines deep convolutional neural networks and generative adversarial networks to normalize the illumination of color or grayscale face images, then train feature extractors and classifiers, and process both frontal and non-frontal face images illumination.The method can be extended to other areas, not only for face image generation.However, it cannot preserve more texture details and has some limitations.Meanwhile, the training model is conducted with well-controlled illumination variations, which can deal with poorly controlled illumination variation to a certain extent, but there are still limitations to the study of other features and geometric structures in realistic and complex environments, etc.It can be further investigated whether the model can work better if the model is trained under complex lighting changes.\cite{bib1}.

When the data set is insufficient, an unsupervised approach can be used for this.For example, for low-light scenes, the unsupervised Aleth-NeRF method is used to learn directly from dark images.The algorithm is mainly a multi-view synthesis method that takes a low-light scene as input and renders a normally illuminated scene.However, a model needs to be trained specifically for different scenes, and also does not handle non-uniform lighting conditions well\cite{bib2}.

Also as far as the results are concerned, images taken in low-light scenes are affected by distracting factors such as blur and noise.For this type of problem, a hybrid architecture based on Retinex theory and Generative Adversarial Network (GAN) can be used to deal with it.For image vision tasks in the dark or under low light conditions, the image is first decomposed into a light image and a reflection image, and then the enhancement part is used to generate a high quality clear image, starting from minimizing the effect of blurring or noise generation.The method introduces Structural Similarity loss to avoid the side effect of blur.But real-life eligible low level and high level images may not be easily acquired and have the shortage of input.Also to maximize the performance of the algorithm, a sufficient size of data set is required.The data obtained after training also has the problem of real-time, which is not enough to meet real-life needs.In general, the algorithm is only from the perspective of solving image blurring and noise, making the impact of these two minimal, other aspects of the problem still exists more, need to further optimize the network structure.\cite{bib3}This class of problems can also be explored by exploring multiple diffusion spaces to estimate the light component, which is used as bright pixels to enhance the shimmering image based on the maximum diffusion value.Generates high-fidelity images without significant distortion, minimizing the problem of noise amplification\cite{bib4}.Later, the conditional diffusion implicit model is utilized in DiFaReli's method (DDIM) to decode the coding of decomposed light.Puntawat Ponglertnapakorn \textit{et~al.} proposed a novel conditioning technique that eases the modeling of the complex interaction between light and geometry by using a rendered shading reference to spatially modulate the DDIM.This method allows for single-view face re-illumination in the wild.However, this method has limitations in eliminating shadows cast by external objects and is susceptible to image ambiguity\cite{bib5}.

In summary,the full objectives of this method are as followed:

$$
\begin{aligned}
L(G, D)= & L_{\text {adversarial }}(G, D)+\lambda_1 \times L_{\text {content }}(\mathrm{G})+\lambda_2 \\
& \times L_{l 1}(\mathrm{G})
\end{aligned}
$$
where $\lambda_1, \lambda_2$ are weight parameters respectively.

$L_{\text {adversarial }}$, $L_{\text {content }}$ and $L_{l 1}$ are as follows:

$$
\begin{aligned}
& L_{\text {adversarial }}(G, D)=E_x[\log D(x)]+E_{G(x)}[\log (1-D(G(x)))] \\
& L_{\text {content }}(G)=\|F(y)-F(G(x))\|_1 \\
& L_{l 1}(G)=\|y-G(x)\|_1
\end{aligned}
$$

where $\mathrm{x}$ denotes input image, whereas $\mathrm{y}$ is the target image, $\mathrm{F}$ means feature extractor.

\subsubsection{Light source movement}A method of generating scenes with a sense of reality from captured object images can be used when the light is moving.On the basis of neural radiation fields (NeRFs), the bulk density of the scene and the radiance of the directional emission are simulated.A representation of each object light transmission is implicitly simulated using illumination and view-related neural networks.This approach can cope with the problem of light movement without retraining the model\cite{bib6}.

\subsubsection{Uneven illumination}For the characteristics of light inhomogeneity in the environment, it is possible to use the light correction network framework, UDoc-GAN, to solve it.The main thing is to convert uncertain normal to abnormal image panning to deterministic image panning with different levels of ambient light for learning guidance.In contrast, Aleth-NeRF cannot handle non-uniform illumination or shadow images.Meanwhile, UDoc-GAN algorithm is more computationally efficient in the inference stage and closer to realistic requirements\cite{bib7}.

\subsubsection{Shadow ray}Jingwang Ling \textit{et~al.} monitored the camera illumination between the scene and multi-view image planes and noticed shadow rays, which led to a new shadow ray supervision scheme. This scheme optimizes the samples and ray positions along the rays. By supervising the shadow rays to achieve controllable illumination, a neural SDF network for single-view scene reproduction under multi-illumination conditions is finally constructed.However,the method is only applicable to point and parallel light sources and has obvious requirements for the position of the light source. The implementation of the method is also based on a simple environment where the scene is not illuminate\cite{bib8}.

\subsubsection{Complex light variation}Also, for uncontrolled complex environment settings from which images are acquired, the NeRF-OSR algorithm enables the generation of new views and new illumination.This is a solution for image generation in complex environments.Solving some fuzzy performance from the perspective of optimizing this algorithm can be an interesting future research direction.For example, resolving inaccuracies in geometric estimation, incorporating more priori knowledge of the outdoor scenes, etc\cite{bib9}.Later, for this problem, Higuera C \textit{et~al.} proposed a solution to the complex problem of light variation by reducing the perceptual differences in vision and using a probabilistic diffusion model to capture light.The method is implemented based on simulated data and can address the limitations of large-scale data.Of course, the method suffers from the problem of computation time, especially in the denoising process which consumes more time\cite{bib10}.This is especially true for reflections in complex environments, for example, with glass and mirrors.YuanChen Guo \textit{et~al.} introduced NeRFReN for simulating scenes with reflections, mainly by dividing the scene into transmission and reflection components and modeling these two components with independent neural radiation fields.This approach has far-reaching implications for further research in scene understanding and neural editing.However, this method does not consider modeling curved reflective surfaces and multiple non-coplanar reflective surfaces\cite{bib11}.

\subsection{Local Illumination}

\subsubsection{Reflectance}Generally speaking, reflected light can be divided into three components, namely ambient reflection, diffuse reflection and specular reflection.The different media materials that cause the reflected light will show different lighting cues in the exposure.An omnidirectional illumination method trains deep neural networks on videos with automatic exposure and white balance to match real images with predicted illumination based on image re-illumination and then regression from the background\cite{bib17}.

The method focuses on minimizing the reconstructed illumination loss function and adding an adversarial loss.And the reconstructed illumination loss and the adversarial loss are as followed:

$$
L_{\mathrm{rec}}=\sum_{b=0}^2 \lambda_b\left\|\hat{M} \odot\left(\Lambda\left(\hat{I}_b\right)^{\frac{1}{\gamma}}-\Lambda\left(I_b\right)\right)\right\|_1 .
$$

In this formulation, the linear rendering of the shear is $\gamma$-encoded with $\gamma$ to match $I$.$\hat{M}$ represents a binary mask.$\lambda_b$ represents an optional weight.

$$
\begin{aligned}
L_{\mathrm{adv}}= & \log D\left(\Lambda\left(I_c\right)\right) \\
& +\log \left(1-D\left(\Lambda\left(\sum_{\theta, \phi} R(\theta, \phi) e^{G(x ; \theta, \phi)}\right)^{\frac{1}{\gamma}}\right)\right)
\end{aligned}
$$

In this formulation, the $D$ represents an auxiliary discriminator network,the $G$ represents the generator,the $x$ represents input image.

Therefore, combining the two yields the following common objectives:
$$
G^*=\arg \min _G \max _D\left(1-\lambda_{\mathrm{rec}}\right) \mathbf{E}\left[L_{\mathrm{adv}}\right]+\lambda_{\mathrm{rec}} \mathbf{E}\left[L_{\mathrm{rec}}\right]
$$

Of course, there are certainly real-life situations where the reflectance is similar.

In illumination variation, there is also a cluster optimization method based on neural reflection field using reflection iteration to solve the problem of similar reflectance of different instances from the perspective of hierarchical clustering.However, there still exists the challenge of facing complex scenarios that do not conform to the unsupervised intrinsic prior, and solutions to such problems need to be proposed\cite{bib18}.

\subsubsection{Radiance}Different mediums have different radiance to light, using a web-based query light integration network on which reflection decomposition is performed.The algorithm captures changing illumination, enabling more accurate new view compositing and re-illumination.
Finally, fast and practical distinguishable rendering areas are implemented.The algorithm can also estimate the shape and BRDF of the objects in the image, which is a point of superiority over other algorithms.However, this method has some limitations in the study of mutual reflection.In particular, an effective treatment of the interactions between all effects could be a future research direction\cite{bib19}.

\section{Application}
\label{sec6}
3D controllable image synthesis has many potential applications in various domains, such as entertainment, industry, and security.

\subsection{Entertainment Application}
\textbf{a)Video games.} 3D image synthesis can create immersive and interactive virtual worlds for gamers to explore and enjoy.  It can also enhance the realism and variety of characters, objects and environments in the game\cite{game_characters,game_characters1}.

\textbf{b)Movies and TV shows.} 3D image synthesis can produce stunning visual effects and animations for movies and TV shows.  It can also enable the creation of digital actors, creatures and scenarios that would be impossible or impractical to film in real life\cite{cartoon1,cartoon}.

\textbf{c)Virtual reality and augmented reality.} 3D image synthesis can generate realistic and immersive virtual experiences for users who wear VR or AR devices.  It can also augment the real world with digital information and graphics that enhance the user's perception and interaction\cite{Metaverse}.

\textbf{d)Art and design.} 3D image synthesis can enable artists and designers to express their creativity and vision in new ways.  It can also facilitate the creation and presentation of 3D artworks, models and prototypes\cite{MappingGothic}.

\subsection{Entertainment Industry}
\textbf{a)Product design and prototyping.} By using 3D image synthesis, designers can visualize and test different aspects of their products, such as shape, color, texture, functionality and performance, before manufacturing them.  This can save time and money, as well as improve the quality and innovation of the products\cite{industry}.

\textbf{b)Training and simulation.}
By using 3D image synthesis, trainers can create realistic and immersive scenarios for workers to practice their skills and learn new procedures.  For example, 3D image synthesis can be used to simulate hazardous environments, such as oil rigs, mines or nuclear plants, where workers can train safely and effectively.

\textbf{c)Inspection and quality control.}
By using 3D image synthesis, inspectors can detect and analyze defects and errors in products or processes, such as cracks, leaks or misalignments.  For example, 3D image synthesis can be used to inspect complex structures, such as bridges, pipelines or aircrafts, where human inspection may be difficult or dangerous\cite{tatsch2023rhino,tian2021hapticenabled}.

\subsection{Entertainment Security}
\textbf{a)Biometric authentication.} 3D image synthesis can be used to generate realistic face images from 3D face scans or facial landmarks, which can be used for identity verification or access control.  For example, Face ID on iPhone uses 3D image synthesis to project infrared dots on the user's face and match them with the stored 3D face model\cite{security,wu2022face}.

\textbf{b)Forensic analysis.} 3D image synthesis can be used to reconstruct crime scenes or evidence from partial or noisy data, such as surveillance videos, witness sketches or DNA samples.  For example, Snapshot DNA Phenotyping uses 3D image synthesis to predict the facial appearance of a person from their DNA\cite{sero2019facial}.

\textbf{c)Counter-terrorism.} 3D image synthesis can be used to detect and prevent potential threats by generating realistic scenarios or simulations based on intelligence data or risk assessment.  For example, the US Department of Defense uses 3D image synthesis to create virtual environments for training and testing purposes.

\textbf{d)Cybersecurity.} 3D image synthesis can be used to protect sensitive data or systems from unauthorized access or manipulation by generating fake or distorted images that can fool attackers or malware.  For example, Adversarial Robustness Toolbox uses 3D image synthesis to generate adversarial examples that can evade or mislead deep learning models\cite{art2018}.

\section{Conclusion}
\label{sec7}
In this paper we have given a comprehensive survey of the emerging progress on 3D controllable image synthesis. We discussed a variety of 3D controllable image synthesis aspects according to their low-level vision cues. The survey reviewed important progress made on 3D datasets, geometrically controllable image synthesis, photometrically controllable image synthesis and related applications. Moreover, the global and local synthesis approaches are separately summarized in each controllable mode to further distinguish diverse synthesis tasks. Our ultimate goal is to provide a useful guide for the researchers and developers who would be interested to synthesizing and editing the image from the low-level 3D prompts. We categorize literatures mainly according to controllable 3D cues since they directly decide our synthesis tasks and abilities. However, there are still other non-rigid 3D cues such as body kinematic joints and elastic shape deformation which are not covered by this survey. In the future, we expect that more explainable controllable cues can be explored from current diffusion and neural radiance fields models by advanced latent decomposition or inverse rendering techniques. Together with the semantic-level controllable image synthesis, the low-level 3D controllable image synthesis and editing can generate more incredible and reliable images in our lives.

\bibliographystyle{IEEEtran}
\bibliography{ref1}

\end{document}